\documentclass[conference]{IEEEtran}
\IEEEoverridecommandlockouts

\usepackage{algorithm}
\usepackage{algpseudocode}
\usepackage{amsfonts}
\usepackage{amsmath}
\usepackage{bm}
\usepackage[T1]{fontenc}
\usepackage{graphicx}
\usepackage{multirow}
\usepackage{subcaption}

\newcommand{\ssep}{\,;\,}
\renewcommand{\exp}[1]{\mathrm{exp}\left(#1\right)}
\renewcommand{\log}[1]{\mathrm{log}\left(#1\right)}
\DeclareMathOperator*{\argmin}{arg\,min}
\newcommand{\pderiv}[2]{\frac{\partial #1}{\partial #2}}

\begin{document}
    \title{Learning Local Feature Aggregation Functions \\ with Backpropagation}
    \author{
        \IEEEauthorblockN{
            Angelos Katharopoulos\footnotemark{*},
            Despoina Paschalidou\footnotemark{*},
            Christos Diou,
            Anastasios Delopoulos
            \thanks{* These two authors contributed equally}
        }
        \IEEEauthorblockA{
            Multimedia Understanding Group\\
            ECE Department, Aristotle University of Thessaloniki, Greece\\
            \{katharas, pdespoin\}@auth.gr; diou@mug.ee.auth.gr; adelo@eng.auth.gr
        }
    }

    \maketitle
    
    \begin{abstract}
    This paper introduces a family of local feature aggregation functions and a
    novel method to estimate their parameters, such that they generate optimal
    representations for classification (or any task that can be expressed as a
    cost function minimization problem). To achieve that, we compose the local
    feature aggregation function with the classifier cost function and we
    backpropagate the gradient of this cost function in order to update the
    local feature aggregation function parameters. Experiments on synthetic
    datasets indicate that our method discovers parameters that model the
    class-relevant information in addition to the local feature space.  Further
    experiments on a variety of motion and visual descriptors, both on image and
    video datasets, show that our method outperforms other state-of-the-art
    local feature aggregation functions, such as Bag of Words, Fisher Vectors
    and VLAD, by a large margin.
    \end{abstract}
    \section{Introduction}\label{sec:intro}

A typical image or video classification pipeline, which uses handcrafted
features, consists of the following components: local feature extraction (e.g.
Improved Dense Trajectories \cite{wang2013action}, SIFT
\cite{lowe2004distinctive}), local feature aggregation (e.g. Bag of Words
\cite{csurka2004visual}, Fisher Vectors \cite{perronnin2010improving}) and
classification of the final aggregated representation. This work focuses on the
second component of the classification pipeline, namely the generation of
discriminative global representations from the local image or video features.

The majority of existing local feature aggregation functions
\cite{csurka2004visual, perronnin2010improving, van2010visual} rely on a visual
codebook learned in an unsupervised manner. For instance, Bag of Words
\cite{csurka2004visual} quantizes every local feature according to a codebook,
most commonly learned with K-Means, and represents the image as a histogram of
codewords. Fisher Vectors \cite{perronnin2010improving}, on the other hand,
capture the average first and second order differences between the local
feature descriptor and the centres of a GMM. Furthermore, the Kernel Codebook
encoding \cite{van2010visual} is analogous to the Bag of Words, with the only
difference that it uses soft assignments, which are functions of the distances
between the local features and the codewords.

There have been several attempts to improve the feature aggregation step by
improving the codebook. For example, the authors of \cite{jurie2005creating}
propose a K-Means alternative that improves modelling of sparse regions of the
local feature space. Other researchers focus on the indirect use of the class
information in order to influence the codebook generation. For instance, in
\cite{lazebnik2009supervised} Lazebnik et al. propose a technique for codebook
learning that aims to minimize the loss of the classification-relevant
information. Finally, in \cite{moosmann2006fast} and \cite{jiu2012supervised}
the authors make direct use of the class labels in order to improve the Bag of
Words representation using a classifier.

In this paper, we define a family of local feature aggregation functions and we
propose a method for the efficient estimation of their parameters
in order to generate optimal representations for classification. In
contrast to former research, our method:
\begin{itemize}
    \item Can be used to estimate any type of parameters and not only codebooks.
    \item Can be used to create representations optimal for any task that can
    be expressed as a differentiable cost function minimization problem, not
    just classification.
\end{itemize}
To demonstrate these properties, we introduce two feature aggregation functions
that outperform state-of-the-art local feature aggregation functions in terms
of classification accuracy in various descriptors for both image and video
datasets.

The rest of the paper is structured as follows. In Section \ref{sec:theory} we
introduce and explain the proposed method. Experimental results are reported in
Section \ref{sec:experiments}, followed by conclusions in Section
\ref{sec:conclusions}.

    \section{Learning Local Feature \\ Aggregation Functions}\label{sec:theory}

Let $F = \{f_1, f_2, \dotsc, f_{N_F}\}$ be the set of $N_F$ local descriptors
extracted from an image or video. In order to derive a global representation
for this feature set, we consider feature aggregation functions that can be
expressed in the form of equation \ref{eq:generic_lfaf}, where $T(\cdot \ssep
\Theta): \mathbb{R}^{D} \mapsto \mathbb{R}^K$ is a differentiable function with
respect to the parameters $\Theta$.

\begin{equation}\label{eq:generic_lfaf}
R(F \ssep \Theta) = \frac{1}{N_F} \sum_{n=1}^{N_F} T(f_n \ssep \Theta)
\end{equation}

By appropriately defining the $T(\cdot \ssep \Theta)$ function, in the above
formulation, we are able to express many local feature aggregation functions.
For instance, the soft-assignment Bag of Words \cite{van2010visual} can be
expressed with the $T(\cdot \ssep \Theta)$ function given in equation
\ref{eq:bow}

\begin{equation}\label{eq:bow}
\begin{aligned}
    T_{BOW}(f_n \ssep C) &=
        \frac{1}{\sum_{k=1}^K D(f_n, C_k)}
        \begin{bmatrix}
            D(f_n, C_1) \\ \vdots \\ D(f_n, C_K)
        \end{bmatrix} \\[0.5em]
\end{aligned}
\end{equation}
where

\begin{equation}\label{eq:D_bow}
\begin{aligned}
    D(f_n, C_k) &= \exp{-\gamma\,(f_n - C_k)^T(f_n - C_k)} \\
\end{aligned}
\end{equation}
is a Gaussian-shaped kernel with Euclidean distance and $C
\in \mathbb{R}^{D \times K}$ is the codebook.

In the following sections, we propose a generic method to estimate the
parameters $\Theta^*$ of the local feature aggregation functions, such that
they generate representations that are optimal for classification. To do that,
we backpropagate the gradient of a classifier's cost function in order to
update the parameters $\Theta$ using gradient descent.

\subsection{Parameter estimation}\label{subsec:optimization}

Most approaches for parameter estimation of local feature aggregation functions
do not take into consideration the subsequent usage of the global feature
representation. For instance, in the case of the classification task, the
extensively used K-Means and GMM methods, ignore the class labels of the
feature vectors in the training set. In this work, we propose a supervised
method for the parameter estimation of any local feature aggregation function
that belongs in the family of functions of equation \ref{eq:generic_lfaf}.
Even though our method can be used for any task that can be expressed as a
differentiable cost function minimization problem, in the rest of this paper
we focus on the classification task.  In particular, we
estimate the values of the parameters $\Theta$ by minimizing the cost function
$J(\cdot)$ of a classifier.

Let $J(x, y \ssep W)$ be the cost function of a classifier with parameters $W$
that aims to predict the class label $y$ from a global feature vector $x$.
Training a classifier is equivalent to finding the $W^* = \argmin_{W}
\frac{1}{N} \sum_{i=1}^N J(x^{(i)}, y^{(i)} \ssep W)$, where $x^{(i)}$ and
$y^{(i)}$ are the $i$-th training sample and its corresponding class label from
a total of $N$ samples. Instead of using traditional clustering methods, such
as K-Means and GMM, to learn the parameters of the feature aggregation
function, we compose $J(\cdot \ssep W)$ with $R(\cdot \ssep \Theta)$. This
allows us to jointly learn a classifier and a feature aggregation function by
solving the optimization problem of equation \ref{eq:optimization_problem}.

\begin{equation}\label{eq:optimization_problem}
W^*, \Theta^* = \argmin_{W, \Theta} \sum_{i=1}^N J\left(R(F^{(i)} \ssep
\Theta), y^{(i)} \ssep W\right)
\end{equation}

Due to the differentiability of $T(\cdot \ssep \Theta)$, a straight-forward way
to solve this optimization problem is to use Stochastic Gradient Descent (SGD).
However, this optimization problem becomes computationally intensive in case of
multimedia and especially for video datasets, due to the large number of local
features $F$ of each video (e.g. more than 20,000 local features in the case of
Improved Dense Trajectories \cite{wang2013action}). In order to address this
problem, we approximate the gradient of $R(\cdot)$ with respect to the $k$-th
parameter $\theta_k$, of equation \ref{eq:generic_gradient}, by using a random
sample of local features, $S_F$, instead of computing the gradient for every
local feature.

\begin{equation}\label{eq:generic_gradient}
\begin{aligned}
\pderiv{J}{\theta_k} &= \pderiv{J}{R(F \ssep \Theta)} \pderiv{R(F \ssep
    \Theta)}{\theta_k} \\
    &= \pderiv{J}{R(F \ssep \Theta)} \frac{1}{N_F}
    \sum_{n=1}^{N_F} \pderiv{T(f_n \ssep \Theta)}{\theta_k} \\
    &\approx \pderiv{J}{R(F \ssep \Theta)} \frac{1}{N_{S_F}}
    \sum_{n \in S_F} \pderiv{T(f_n \ssep \Theta)}{\theta_k}
\end{aligned}
\end{equation}

Empirical results indicate that this approximation has similar effects to the
stochastic gradient approximation of SGD, namely efficiency and robustness.

\subsection{Aggregation functions}\label{subsec:implementations}

In this section, we make use of the previous analysis in order to create two
local feature aggregation functions that outperform other state-of-the-art
methods such as Bag of Words \cite{csurka2004visual} and Fisher Vectors
\cite{perronnin2010improving} on a variety of descriptors, as shown in the
Experiments section \ref{sec:experiments}.

Firstly, we consider the representation $R_{1}(\cdot)$, which is a
generalization of the soft-assignment Bag of Words and employs the encoding
function $T_{1}(\cdot)$ of equation \ref{eq:t1}

\begin{equation}\label{eq:t1}
\begin{aligned}
    T_{1}(f_n \ssep C, \Sigma) &=
        \frac{1}{Z(f_n, C, \Sigma)}
        \begin{bmatrix}
            D(f_n, C_1, \Sigma_1) \\ \vdots \\ D(f_n, C_K, \Sigma_K)
        \end{bmatrix} \\[0.5em]
\end{aligned}
\end{equation}
where
\begin{equation}\label{eq:D}
\begin{aligned}
    D(f_n, C_k, \Sigma_k) &= \exp{-\gamma\,(f_n - C_k)^T\Sigma_k^{-1}(f_n - C_k)} \\
\end{aligned}
\end{equation}
and
\begin{equation}\label{eq:Z}
\begin{aligned}
    Z(f_n, C, \Sigma) &= \sum_{k=1}^K D(f_n, C_k, \Sigma_k) \\
\end{aligned}
\end{equation}
involve the codebook $C_k$ and the diagonal covariance matrix $\Sigma_k$ used
to compute the Mahalanobis distance between the $n$-th local feature and the
$k$-th codeword.

On the other hand, we consider the representation $R_{2}(\cdot)$, produced by
the encoding function $T_{2}(\cdot)$ of equation \ref{eq:t2}, which is exactly
the soft-assignment Vector of Locally Aggregated Descriptors (VLAD)
\cite{jegou2010aggregating} and thus the dimensionality of the resulting
representation is $D \times K$ because $f_n - C_k$ is a vector of size $D$.

\begin{equation}\label{eq:t2}
\begin{aligned}
    T_{2}(f_n \ssep C, \Sigma) &=
        \frac{1}{Z(f_n, C, \Sigma)}
        \begin{bmatrix}
            D(f_n, C_1, \Sigma_1)(f_n - C_1) \\
            \vdots \\
            D(f_n, C_K, \Sigma_K)(f_n - C_K)
        \end{bmatrix} \\[0.5em]
\end{aligned}
\end{equation}

In order to compute the optimal parameters $C$ and $\Sigma$ of the local
feature aggregation functions, we optimize equation
\ref{eq:optimization_problem} using a Logistic Regression classifier with a
cross-entropy loss according to equation \ref{eq:logistic_regression}.  While
linear classifiers are very efficient, non-linear classifiers tend to yield
better classification results, especially in the case of Bag of Words
\cite{zhang2007local}. Therefore, we decided to adopt an approximate feature
map of $\chi^2$ \cite{vedaldi2012efficient} that is used in combination with
$T_{1}(\cdot)$ and Logistic Regression to retain both the training efficiency
of a linear classifier and the classification accuracy of a non-linear
classifier.

\begin{equation}\label{eq:logistic_regression}
J(x, y \ssep W) = -\log{\frac{\exp{W_y^T x}}
    {\sum_{\hat{y}} \exp{W_{\hat{y}}^T x}}}
\end{equation}

We could have used any classifier whose training is equivalent to minimizing a
differentiable cost function, such as Neural Networks. Nevertheless, we use
Logistic Regression and a $\chi^2$ feature map in order to fairly compare our
method to existing feature aggregation functions.

\subsection{Training procedure}\label{subsec:training}

In Algorithm \ref{alg:training}, we present the training procedure for the
feature aggregation functions introduced in Section
\ref{subsec:implementations}. The training process consists of three main
parts, the initialization step, the optimization step and the classifier
fine-tuning step.

Regarding the initialization, we have experimented with three methods to
initialize the codebook $C$ and the covariance matrices $\Sigma$. In
particular, we used: 
\begin{itemize}
    \item Random sampling from the set of local features to initialize
          the codebook and the identity matrix to initialize the covariance
          matrices.
    \item K-Means clustering to initialize the codebook and the identity matrix
          to initialize the covariance matrices.
    \item GMM clustering to initialize both the codebook and the covariance
          matrices
\end{itemize}
The proposed method can be used in combination to any of the aforementioned
initializations. However, we empirically observe that when initialized
with K-Means it results in a smoother parameter space, hence it is easier to
choose a suitable value for the SGD learning rate.  Finally, the reason for
adding the classifier fine-tuning step emerged from the need to alleviate the
effects of gradient noise, produced by the sampling of local features in
equation \ref{eq:generic_gradient}.

\begin{algorithm}
\caption{Procedure to learn the parameters of a local feature aggregation
function}\label{alg:training}
\begin{algorithmic}
    \Procedure{TrainAggFun}{$F, y$}
        \State \# Parameter initialization
        \If{initialize with K-Means}
            \State $C_0 \gets KMeans(F)$
            \State $\Sigma_0 \gets \mathbf{I}$
        \Else
            \State $C_0, \Sigma_0 \gets GMM(F)$
        \EndIf
        \State $W_0 \gets \argmin_W \sum_{i=1}^N J\left(R(F^{(i)} \ssep
                C_0, \Sigma_0), y^{(i)} \ssep W\right)$
        \State \# Core training
        \State $t \gets 0$
        \Repeat
            \State $i \sim \text{DiscreteUniform}(1, N)$
            \State Sample $\hat{F}^{(i)}$ from $F^{(i)}$
            \State $W_{t+1} \gets \mathrm{SGD}(\nabla_{W_t} J(R(\hat{F}^{(i)} \ssep C_t, \Sigma_t), y^{(i)} \ssep W_t))$
            \State $C_{t+1} \gets \mathrm{SGD}(\nabla_{C_t} J(R(\hat{F}^{(i)} \ssep C_t, \Sigma_t), y^{(i)} \ssep W_t))$
            \State $\Sigma_{t+1} \gets \mathrm{SGD}(\nabla_{\Sigma_t} J(R(\hat{F}^{(i)} \ssep C_t, \Sigma_t), y^{(i)} \ssep W_t))$
            \State $t \gets t + 1$
        \Until{$t \geq \text{specific number of mini-batches}$}
        \State \# Classifier fine tuning
        \State $C^* \gets C_t$
        \State $\Sigma^* \gets \Sigma_t$
        \State $W^* \gets \argmin_W \sum_{i=1}^N J\left(R(F^{(i)} \ssep
                C^*, \Sigma^*), y^{(i)} \ssep W_t\right)$
    \EndProcedure
\end{algorithmic}
\end{algorithm}

    \section{Experiments}\label{sec:experiments}

This section presents an experimental evaluation of the proposed method on real
and artificial datasets in order to assess its effectiveness and provide
insights into the resulting feature aggregation functions. In particular, we
have conducted experiments on the CIFAR-10 \cite{krizhevsky2009learning} image
classification dataset and the UCF-11 (YouTube) Action dataset
\cite{liu2009recognizing}. In case of CIFAR-10, we have extracted local
features with a pre-trained deep convolutional neural network.  Specifically,
we have used the \textit{conv3\_3} layer from VGG-16 architecture
\cite{DBLP:journals/corr/SimonyanZ14a}, pre-trained on Imagenet, which results
in 25 local features in $\mathbb{R}^{256}$ for each image. In addition, in case
of the video data, we have extracted Improved Dense Trajectories
\cite{wang2013action}, after removing videos that have less than 15 frames,
which results on an average of approximately 22,000 local features per video.

In Section \ref{subsec:synthetic_dataset}, we present a comparative evaluation
of the discovered codewords in two synthetic datasets, in order to acquire a
better understanding of the way our method chooses the codebook, compared to
unsupervised methods. Subsequently, in Section \ref{subsec:training_evolution},
we present the classification accuracy of various representations on CIFAR-10,
with respect to the training epochs, and compare it to the corresponding
results using Bag of Words. Finally in Section \ref{subsec:classification}, we
compare the proposed method on CIFAR-10 and UCF-11 with respect to the
classification accuracy to Fisher Vectors, Bag of Words and VLAD on a variety
of descriptors.

\subsection{Synthetic dataset}\label{subsec:synthetic_dataset}

\begin{figure*}
    \centering
    \begin{subfigure}[t]{1\textwidth}
        \centering
        \includegraphics[width=0.8\linewidth]{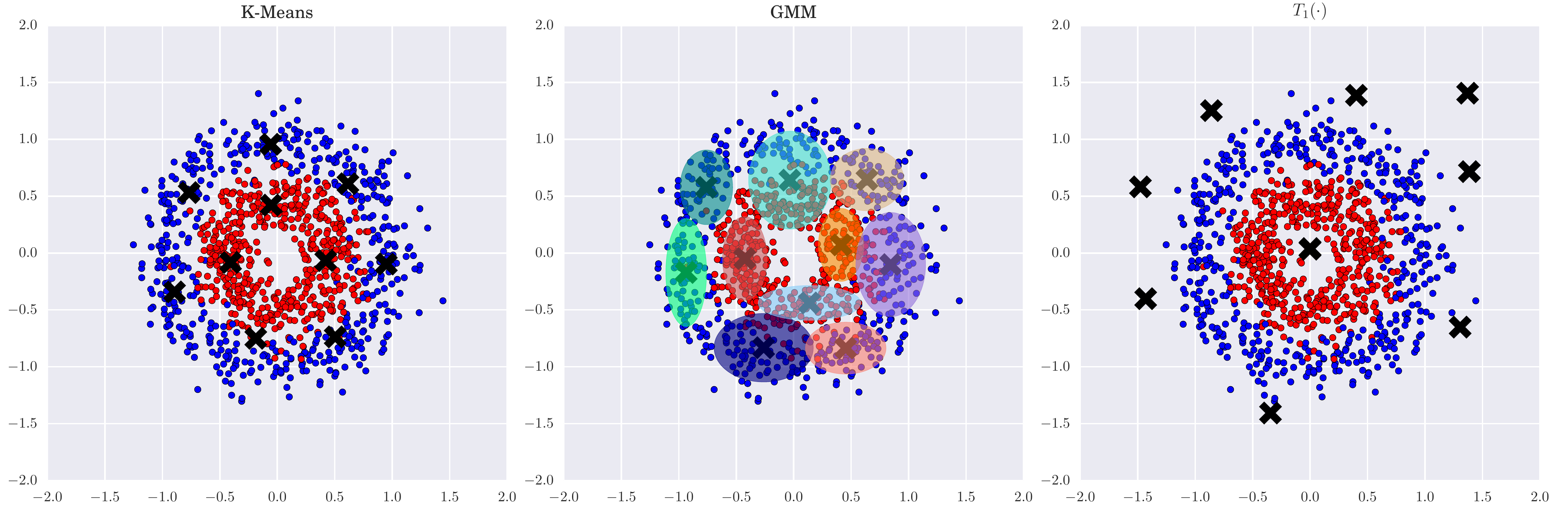}
        \caption{Concentric}\label{subfig:concentric}
    \end{subfigure}
    \begin{subfigure}[t]{1\textwidth}
        \centering
        \includegraphics[width=0.8\linewidth]{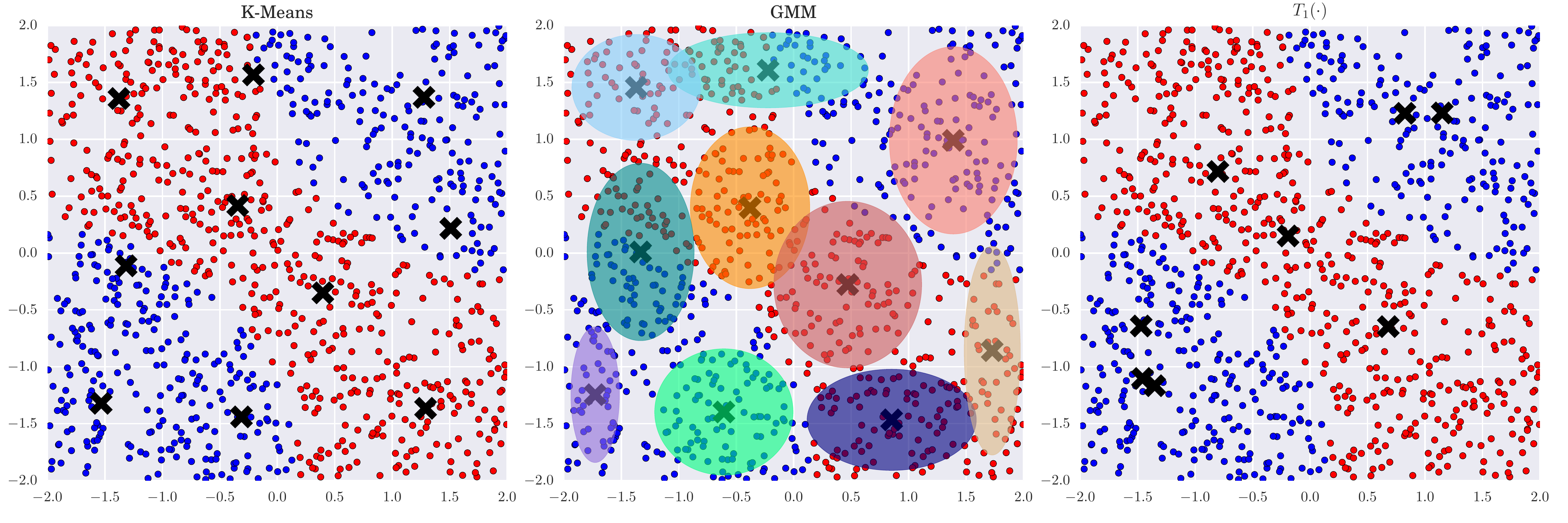}
        \caption{Non-linear (XOR)}\label{subfig:nonlinear}
    \end{subfigure}
    \caption{Generated codebooks from synthetic data by K-Means, GMM and our method.
    The generated codewords are drawn with black crosses, while the dots are the local
    features from both classes (either blue or red).}
    \label{fig:artificial_data}
\end{figure*}

Figure \ref{fig:artificial_data} compares the generated codebooks by K-Means,
GMM and the proposed method on two artificial two-class datasets. In both
cases, we generate and visualize 10 codewords, especially in the case of GMM we
visualize additionally the covariance matrices. For our method, we use the
$T_1(\cdot)$ feature aggregation function, from equation \ref{eq:t1}, to learn
the codebook with the covariance matrices being fixed and equal to the identity
matrix. 

In contrast to K-Means and GMM, our method focuses on generating
representations that can be separated by the classifier without necessarily
retaining the structural information of the local features. It only suffices to
observe Figure \ref{subfig:concentric} to note that K-Means and GMM do not
respect the circular class boundary, while our method focuses mainly on
generating a linearly separable representation. In addition, owing to the fact
that our method does not try to describe the local features it results in a
more separable representation with a smaller codebook. For instance, it only
requires a single codeword to successfully separate the concentric dataset of
Figure \ref{subfig:concentric}.

\subsection{Training evolution}\label{subsec:training_evolution}

\begin{figure}
    \centering
    \includegraphics[width=0.49\textwidth]{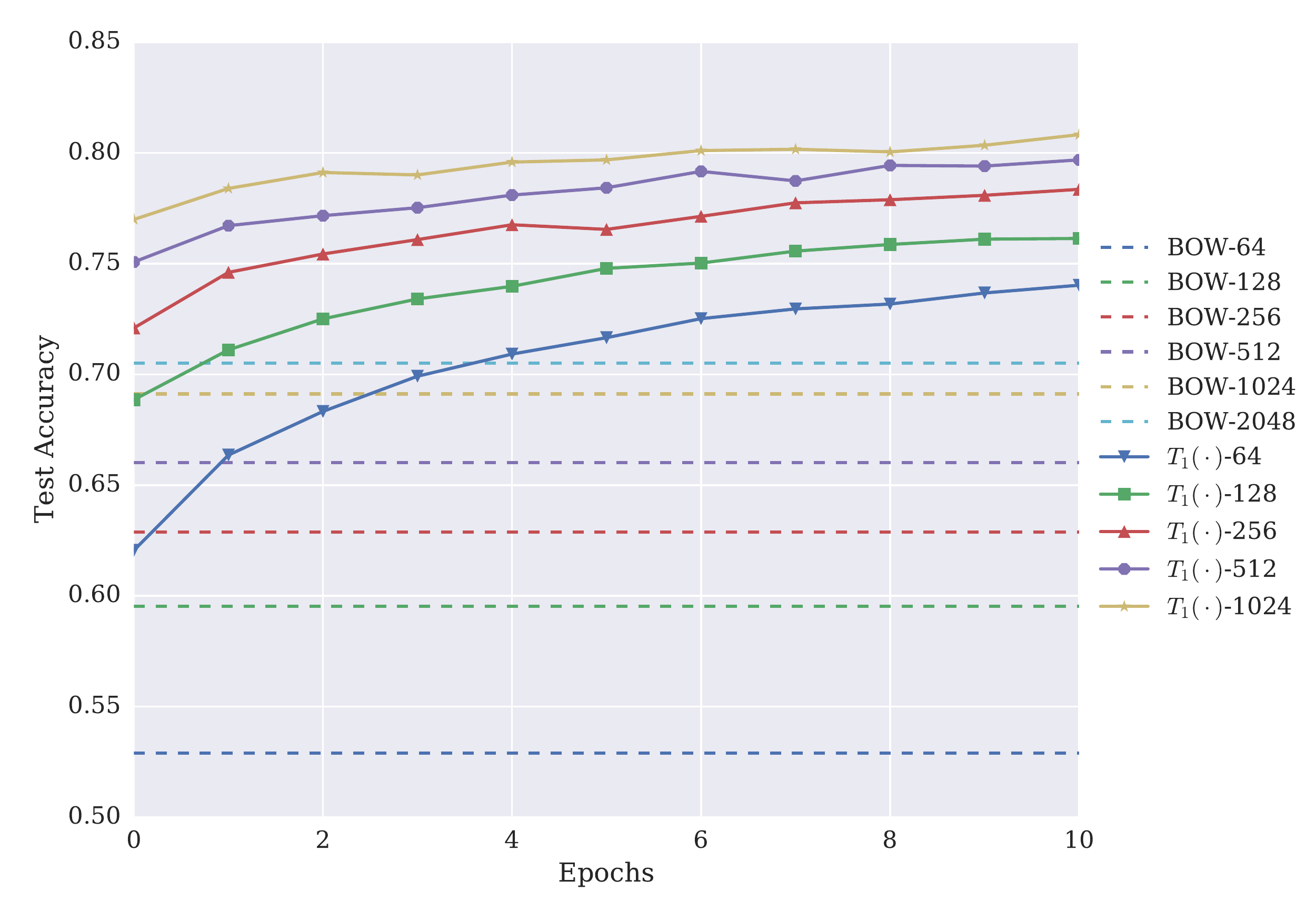}
    \caption{Classification accuracy on the test set with respect to the training
    epochs for various representation sizes on CIFAR-10 with features from VGG16
    conv3\_3}
    \label{fig:train_evolution}
\end{figure}

For this experiment, we generate codebooks using K-Means of sizes $\{64, 128,
256, 512, 1024, 2048\}$, which we subsequently use to create the corresponding
Bag of Words representations. To classify the produced representations, we
train a linear SVM with a $\chi^2$ feature map. Moreover, we use the
$T_1(\cdot)$ feature aggregation function, of equation \ref{eq:t1}, with
Logistic Regression, a $\chi^2$ feature map and K-Means as an initialization
method according to Algorithm \ref{alg:training}. In order to select a value
for the hyper-parameter $\gamma$ of the $T_1(\cdot)$ function, we perform
cross-validation.

By observing Figure \ref{fig:train_evolution}, we conclude that the proposed
method produces discriminative representations even with a small number of
dimensions.  In particular, it outperforms Bag of Words with 2048 dimensions by
almost $4$ percentage points with only 64 dimensions. Furthermore, we also
notice that our method considerably improves the representation during the
first epochs, thus we conclude that it can be used to fine-tune any
differentiable feature aggregation function (e.g. Fisher Vectors) with little
computational effort. Finally, we anticipate that increasing the number of
training epochs will further increase the classification accuracy.

\subsection{Classification results}\label{subsec:classification}

In the current experiment, we assess the discriminativeness of the produced
representations by evaluating their classification performance on a variety of
descriptors and comparing it to several state-of-the-art feature aggregation
methods. In the case of CIFAR-10, we use the provided train-test split while
for UCF-11, we create three random 60/40 train-test splits and report both the
mean classification accuracy and the standard error of the mean. Table
\ref{tab:classification} summarizes the results. The experimental setup for
CIFAR-10 is analysed in Section \ref{subsec:training_evolution}.  Regarding
UCF-11, we generate codebooks of sizes $\{1024, 2048\}$ using K-Means, both to
create Bag of Words representations and to initialize the codebooks for the
$T_1(\cdot)$ function. In addition, we train a GMM with 64 Gaussians to
generate Fisher Vectors representations and again K-Means with 64 centroids to
generate  VLAD and initialize $T_2(\cdot)$.

For both datasets, we train an SVM with a $\chi^2$ feature map for Bag of Words
and $T_1(\cdot)$ and a linear SVM for the rest of the local feature aggregation
functions in Table \ref{tab:classification}.  Moreover, in case of CIFAR-10,
$T_1(\cdot)$ is trained for only 10 epochs, while for UCF-11, both $T_1(\cdot)$
and $T_2(\cdot)$ are trained for 30 epochs. In the conducted experiments, we
have observed that both $T_1(\cdot)$ and $T_2(\cdot)$ are very sensitive with
respect to the hyper-parameter $\gamma$, which must be carefully selected using
a validation set or cross-validation. In particular, the reported results are
generated using $\gamma=70$ for UCF-11 ``idt\_traj'', $\gamma=50$ for UCF-11
``idt\_hof'' and $\gamma=5 \times 10^{-8}$ for CIFAR-10. The large differences
in the range of $\gamma$ make intuitive sense upon observing the distribution
of the pairwise distances of the local features.

Furthermore, we
additionally report the classification accuracy attained by $T_1(\cdot)$ and
$T_2(\cdot)$, without learning the parameters using the proposed method; the
results are reported in Table \ref{tab:classification} as ``initial''. This
allows us to quantify the improvement in terms of classification accuracy
achieved using the proposed method. In particular, we observe an average
improvement of approximately 3.5 percentage points in all cases.

\bgroup
\def\arraystretch{1.2}
\begin{table}
\centering
\caption{Comparison of $T_1(\cdot)$ and $T_2(\cdot)$ with Bag Of Words (BOW),
VLAD and Fisher Vectors (FV)}
\label{tab:classification}
\begin{tabular}{|l|c||c|c|}
\cline{2-4}
\multicolumn{1}{c|}{} & CIFAR-10 & \multicolumn{2}{|c|}{UCF-11} \\
\cline{2-4}
\multicolumn{1}{c|}{} & conv3\_3 & idt\_hof & idt\_traj \\
\noalign{ \hrule height 0.15em }
BOW-1024 & 69.12\% & 89.72\% +/- 0.50 & 83.88\% +/- 0.39 \\
\hline
BOW-2048 & 70.50\% & 91.03\% +/- 0.35 & 85.65\% +/- 0.53 \\
\hline
$T_1(\cdot)$-1024 initial & 76.85\% & 88.41\% +/- 0.46 & 82.84\% +/- 0.80 \\
\hline
$T_1(\cdot)$-1024 & 80.87\% & 92.23\% +/- 0.37 & 86.90\% +/- 0.63 \\
\hline
$T_1(\cdot)$-2048 initial &  78.52\% & 90.24\% +/- 0.24 & 83.78\% +/- 0.90 \\
\hline
$T_1(\cdot)$-2048 & \textbf{81.12\%} & \textbf{93.00\% +/- 0.30} & \textbf{87.01\% +/- 0.48} \\
\noalign{ \hrule height 0.15em }
VLAD-64 & - & 90.25\% +/- 0.33 & 78.71\% +/- 0.94 \\
\hline
FV-64 & - & 90.55\% +/- 0.26 & 78.92\% +/- 0.21 \\
\hline
$T_2(\cdot)$-64 initial  & - & 90.92\% +/- 0.15 & 79.96\% +/- 0.46 \\
\hline
$T_2(\cdot)$-64  & - & \textbf{91.08\% +/- 0.26} & \textbf{83.82\% +/- 0.34} \\
\hline
\end{tabular}
\end{table}
\egroup

    \section{Conclusions}\label{sec:conclusions}

We have introduced a new method to learn the parameters of a family of local
feature aggregation functions through optimization, which can be used to learn
any type of parameters and is not limited to codebooks.  Furthermore, it can be
used to generate an optimal representation for any task that can be expressed
as a cost function minimization problem. In particular, in the conducted
experiments, we have demonstrated the effectiveness of the proposed method in
the classification task.  We observed that the proposed local feature
\vfill\eject
\noindent aggregation functions outperform Bag of Words, Fisher Vectors and
VLAD in a variety of descriptors on image and video data.

Our method opens up a multitude of new research directions. Initially, we could
use the proposed method to learn extra parameters, such as $\gamma$, in order
to further improve the generated representation. Moreover, it would be
interesting to conduct experiments on other large-scale video classification
datasets, such as UCF101 \cite{DBLP:journals/corr/abs-1212-0402} and compare
the performance of our method to state-of-the-art Neural Network architectures,
such as the hybrid deep learning framework, as it was introduced in
\cite{wu2015modeling}.  Finally, we can explore the use of the proposed method
for the generation of optimal representations for other types of tasks, such as
regression or ranking.

    \bibliographystyle{abbrv}
    \bibliography{references}
\end{document}